\documentclass[letterpaper]{article} 
\usepackage{aaai24}  
\usepackage{times}  
\usepackage{helvet}  
\usepackage{courier}  
\usepackage[hyphens]{url}  
\usepackage{graphicx} 
\usepackage{natbib}  
\usepackage{caption} 
\usepackage{algorithm}
\usepackage{algorithmic}

\def\eg{\emph{e.g.}} 
\def\ie{\emph{i.e.}} 
 
\def\etc{\emph{etc}} \def\vs{\emph{vs.}}

\usepackage{graphicx}
\usepackage{amsmath}
\usepackage{amssymb}
\usepackage{booktabs}
\usepackage{multirow}
\usepackage{enumitem}
\usepackage{multirow}
\usepackage{float}
\usepackage{colortbl}
\usepackage{color} 
\usepackage{makecell}

\usepackage{newfloat}
\usepackage{listings}
\DeclareCaptionStyle{ruled}{labelfont=normalfont,labelsep=colon,strut=off} 
\floatstyle{ruled}
\newfloat{listing}{tb}{lst}{}
\floatname{listing}{Listing}
\title{SSMG: Spatial-Semantic Map Guided Diffusion Model for Free-Form Layout-to-Image Generation}
\author{
   Chengyou Jia\textsuperscript{\rm 1}\thanks{This work was completed during the internship at SGIT AI Lab, State Grid Corporation of China.},\;  Minnan Luo\textsuperscript{\rm 1}\thanks{Corresponding author.},\; Zhuohang Dang\textsuperscript{\rm 1},\; Guang Dai\textsuperscript{\rm 2,3},\; Xiaojun Chang\textsuperscript{\rm 4,5}, \\ Mengmeng Wang\textsuperscript{\rm 6,2} , \; Jingdong Wang\textsuperscript{\rm 7}
}
\affiliations{
    \textsuperscript{\rm 1}School of Computer Science and Technology, MOEKLINNS Lab, Xi’an Jiaotong University \; \\ \textsuperscript{\rm 2}SGIT AI Lab \\
    \textsuperscript{\rm 3}State Grid Corporation of China \\
    \textsuperscript{\rm 4}University of Technology Sydney  \\ 
    \textsuperscript{\rm 5}Mohamed bin Zayed University of Artificial Intelligence \\
    \textsuperscript{\rm 6}Zhejiang University \\
    \textsuperscript{\rm 7}Baidu Inc\\
\{cp3jia, dangzhuohang\}@stu.xjtu.edu.cn,  minnluo@xjtu.edu.cn,
 \{guang.gdai, cxj273\}@gmail.com,  mengmengwang@zju.edu.cn,  wangjingdong@outlook.com
}

\usepackage{bibentry}

\begin{document}

\maketitle

\begin{abstract}
Despite significant progress in Text-to-Image (T2I) generative models, even lengthy and complex text descriptions still struggle to convey detailed controls. In contrast, Layout-to-Image (L2I) generation, aiming to generate realistic and complex scene images from user-specified layouts, has risen to prominence. However, existing methods transform layout information into tokens or RGB images for conditional control in the generative process, leading to insufficient spatial and semantic controllability of individual instances. To address these limitations, we propose a novel Spatial-Semantic Map Guided (SSMG) diffusion model that adopts the feature map, derived from the layout, as guidance. Owing to rich spatial and semantic information encapsulated in well-designed feature maps, SSMG achieves superior generation quality with sufficient spatial and semantic controllability compared to previous works. Additionally, we propose the Relation-Sensitive Attention (RSA) and Location-Sensitive Attention (LSA) mechanisms. The former aims to model the relationships among multiple objects within scenes while the latter is designed to heighten the model's sensitivity to the spatial information embedded in the guidance. Extensive experiments demonstrate that SSMG achieves highly promising results, setting a new state-of-the-art across a range of metrics encompassing fidelity, diversity, and controllability. 
\end{abstract}

\begin{figure*}[t]
  \centering
  \includegraphics[width=0.85\textwidth]{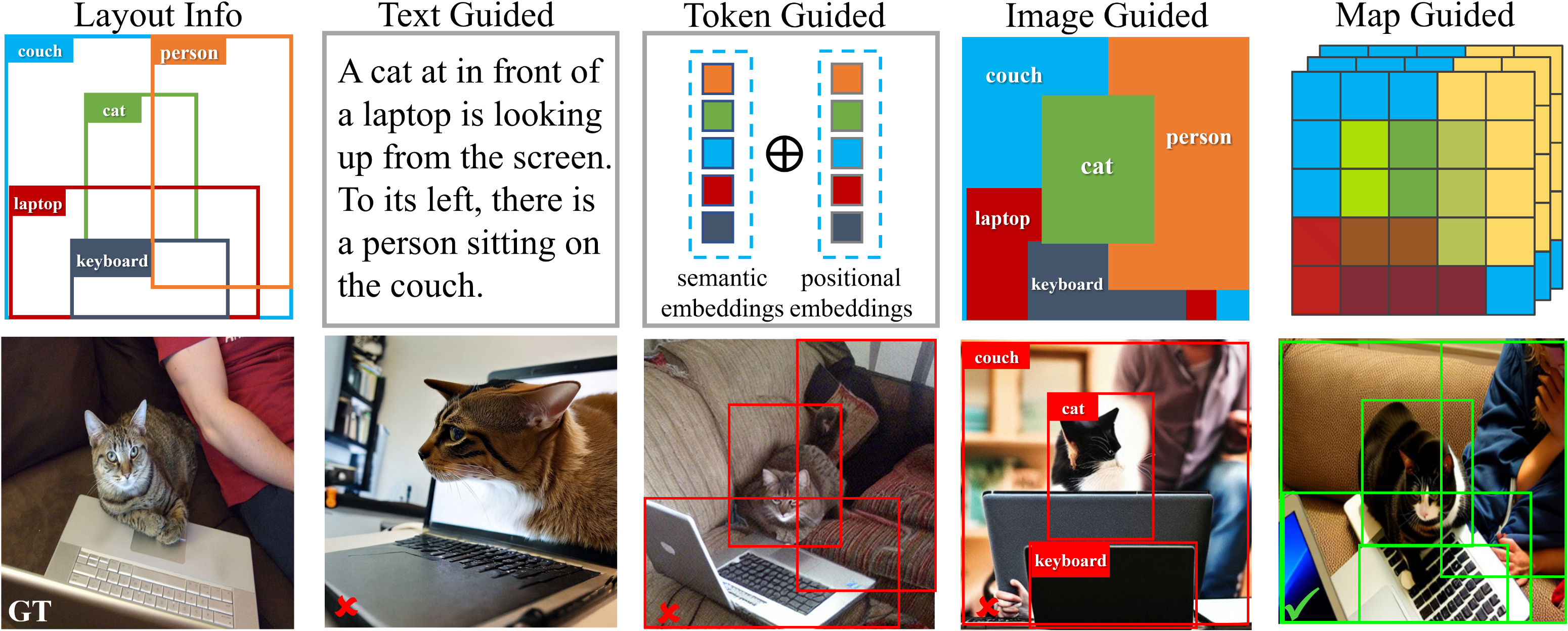}
  \caption{A comparison of image generation methods with different guidance. 
  In contrast to the prevalent token-guided or image-guided methods, our map-guided method excels in providing superior control over both the spatial arrangement and semantic details of individual instances, thereby leading to higher-quality and more appropriate results.} 
  \label{fig:intro}
\end{figure*}

\section{Introduction}

Despite notable advancements in Text-to-Image (T2I) generative models \cite{nichol2021glide,saharia2022photorealistic}, text descriptions often struggle to adequately convey detailed controls, even when composed of lengthy and complex texts (as shown ``Text Guided'' in Figure \ref{fig:intro}). 
In contrast, methods of Layout-to-Image (L2I) generation have risen to prominence, enabling the rendering of realistic and complex scene images from user-specified layouts. Owing to its fine-grained control, L2I generation possesses enormous potential for a wide range of display applications. These range from enhancing the user experience in interactive design to revolutionizing content creation in film and animation \cite{zhu2016generative,isola2017image,10081412}.  


Recent methods adapt large-scale pre-trained diffusion models \cite{rombach2022high} for the L2I generation, demonstrating superior performance compared to GAN-based approaches \cite{park2019semantic,richardson2021encoding,lee2020maskgan}.  
These methods \cite{li2023gligen,yang2023reco,zheng2023layoutdiffusion,cite} can be broadly classified into two types, token-guided and image-guided. 
Token-guided methods, as illustrated in Figure \ref{fig:intro}, tokenize the layout of spatial and semantic information to embeddings, which are then integrated into T2I models via an attention mechanism. 
Semantic information ``classes'' is typically managed by a text encoder, \eg, CLIP \cite{radford2021learning}, while spatial position information is handled by various tokenization methods, \eg, Fourier embedding in GLIGEN \cite{li2023gligen} or learnable matrixes in ReCO \cite{yang2023reco}. 
However, the tokenization of spatial information, akin to textualization, fails to effectively harness the inherent 2D spatial structure of the layout. Consequently, as demonstrated in Figure \ref{fig:intro}, it struggles to achieve fine-grained spatial control at the pixel level, leading to issues such as imprecise object boundaries and loss of objects. In contrast, ControlNet \cite{cite}, a representative image-guided work, incorporates visual conditions, \eg, the layout of boxes, into frozen T2I diffusion models to enable additional condition signals. ControlNet duplicates the weights of the diffusion model into a ``trainable copy'' and a ``locked copy'', with the trainable copy being trained to learn these visual conditions. 
This approach provides more accurate spatial controllability. Nevertheless, the semantic control in ControlNet is solely derived from global image captions, resulting in a lack of control over the detailed semantics of individual instances. As shown in Figure \ref{fig:intro}, the objects within the box suffer from severe semantic ambiguity, resulting in erroneous or blurred objects.

Considering the aforementioned limitations, a pressing question arises: can a novel guiding strategy be developed that overcomes the shortcomings of both token-guided and image-guided strategies, thereby offering precise control over both the semantics and the spatial layout of the generated images? In response to this contemplation, we argue that a map-guided strategy could serve as an effective solution. 
On the one hand, the 2D feature map naturally inherits the spatial structure of layouts, which preserves the advantage of visual conditions as exemplified in ControlNet. 
On the other hand, feature maps offer a richer semantic dimension than the RGB space of image guidance, \eg, $H \times W \times C$, where $C$$\gg$$3$. This enriched semantic dimension allows for a more detailed content representation at each spatial pixel. As a result, it enables the effective control of positions and semantics of individual instances, achieving enhanced spatial accuracy and semantic richness.

In this paper, we present a novel Spatial-Semantic Map Guided (SSMG) diffusion model that adopts the feature map generated from layout as guidance. Specifically, we first initialize the spatial-semantic map based on the given layout info. By fully embracing the spatial structure inherent in the layout, we populate the corresponding position in the map with semantics learned in the textual encoder. Our initialization not only ensures the complete preservation of spatial information but also infuses the feature map with rich semantic content. Then, considering that the initial map processes each instance independently, we further propose Relation-Sensitive Attention (RSA) to establish the relationships among instances in the scene, as well as the relationship of each instance with the overall scene. RSA allows the spatial and semantic information of each instance to cross-reference all other instance or scene information, thereby providing a more nuanced understanding of the scene context. Finally, with the enhanced spatial-semantic map, we integrate it into the conditional generation process through the proposed Location-Sensitive Attention (LSA). This strategy empowers the model to warp the noise image features at each position according to the spatial-semantic map, heightening the model’s sensitivity to the embedded spatial information. 
In such a way, our SSMG effectively enhances the model's controllability over both semantics and the spatiality of generated images. Moreover, our approach serves as a significant extension to traditional T2I methods. SSMG not only permits free-form textual descriptions for each instance but also supports a multitude of layout positional representations, transcending the limitations of bounding boxes. These advantages provide ample 
adaptability and flexibility for L2I generation, rendering it convenient for wider applications.


Experiments conducted on benchmark datasets demonstrate that our SSMG achieves highly promising results, setting a new state-of-the-art across a range of metrics encompassing fidelity, diversity, and controllability. Notably, SSMG verifies the superiority of the map-guided strategy in layout controllability, significantly improving previous state-of-the-arts YOLO scores from $30.5$ to $37.6$ on the COCO dataset. We also provide sufficient representative cases to highlight the distinctive features of our method.



\section{Related Work}
\paragraph{Layout-to-Image Generation}

Layout-to-image generation with given bounding boxes can be viewed as 
a reverse process of object detection. 
Early GAN-based methods adopt the encoder-decoder architecture to transform boxes into images. LAMA \cite{LAMA}, LostGANs \cite{sun2021learning}, and Context L2I \cite{he2021context} encoded the layout of boxes as a style feature, subsequently feeding this into an adaptive normalization layer. Adding to these GAN-based methods, Taming \cite{esser2021taming} and TwFA \cite{yang2022modeling} encode layout information as inputs to a transformer. They then employed an auto-regressive (AR) approach to predict the latent visual codes. Recently, diffusion-based methods \cite{ho2020denoising, songdenoising, rombach2022high} show promising results in L2I. LDM \cite{rombach2022high} employed a T2I model for box-to-image transfer, encoding layouts using a BERT model. Beyond that, \cite{li2023gligen,yang2023reco,zheng2023layoutdiffusion} tokenized bounding boxes into embeddings by Fourier embedding \cite{mildenhall2021nerf} or new trainable layers. These embeddings are then injected into pre-trained T2I models, facilitating the generation of scene images with positional information. 

Note that existing L2I methods are limited to dealing with the form of bounding boxes. Other forms of layout, \eg, key points, semantic masks in Figure \ref{fig:free_layout}, require specialized methods \cite{park2019semantic,richardson2021encoding,lee2020maskgan,xue2023freestyle}. 
In contrast, this paper introduces a novel map-guided approach that transforms given layout info into feature maps, which subsequently guide the model's learning process. Crucially, our approach is versatile and can be applied to a variety of forms of layout.



\paragraph{Diffusion Model} Deep diffusion-based generative models \cite{ho2020denoising,song2021scorebased,dhariwal2021diffusion} have showcased their exceptional ability to generate high-quality and diverse samples. Building upon these works, LDM \cite{rombach2022high} utilized VQ-VAE \cite{oord2017neural} to encode images into latent codes of smaller resolution, thereby reducing the computational overhead required for training a diffusion model. LDM also achieved impressive results in T2I generation by encoding text inputs into latent vectors, similar to approaches, GLIDE \cite{nichol2021glide}, DALL-E 2 \cite{ramesh2022hierarchical}, and Imagen \cite{saharia2022photorealistic}.

Beyond text-guided generation, ControlNet \cite{cite} enabled additional image-guided conditions into frozen T2I diffusion models, \eg, sketch, segmentation masks, canny edge, \etc. It duplicates the weights of a large diffusion model into a ``trainable copy'' and a ``locked copy'' and the trainable copy is trained to learn these conditional signals. 
While ControlNet has demonstrated a commendable ability to control the shape or position of generated objects, it falls short in providing detailed semantic descriptions for individual instances, thereby failing to address L2I effectively. 
To overcome this, we introduce a novel map-guided approach that incorporates robust semantic and spatial awareness to enhance the model's controllability.

\section{Methodology}
\subsection{Problem Definition}
In Layout-to-Image (L2I) generation, we are presented with a set of instance entities $\boldsymbol{e} = \{(t_k,p_k)|k=1,\cdots,n\}$, where each entity $\boldsymbol{e}_k$ consists of textual description $t_k$ and positional information $p_i$. In addition to $\boldsymbol{e}$, a global textual description $t_{global}$ is provided, serving as guidance for the overall style and content of the generated image. 
The L2I model can be conceptualized as a function $f(\boldsymbol{e},t_{global}) = x$, aiming to generate a realistic image $x$ from the input layout information of entities $\boldsymbol{e}$ and the global description $t_{global}$. 

\begin{figure*}[t]
  \centering
  \includegraphics[width=0.85\textwidth]{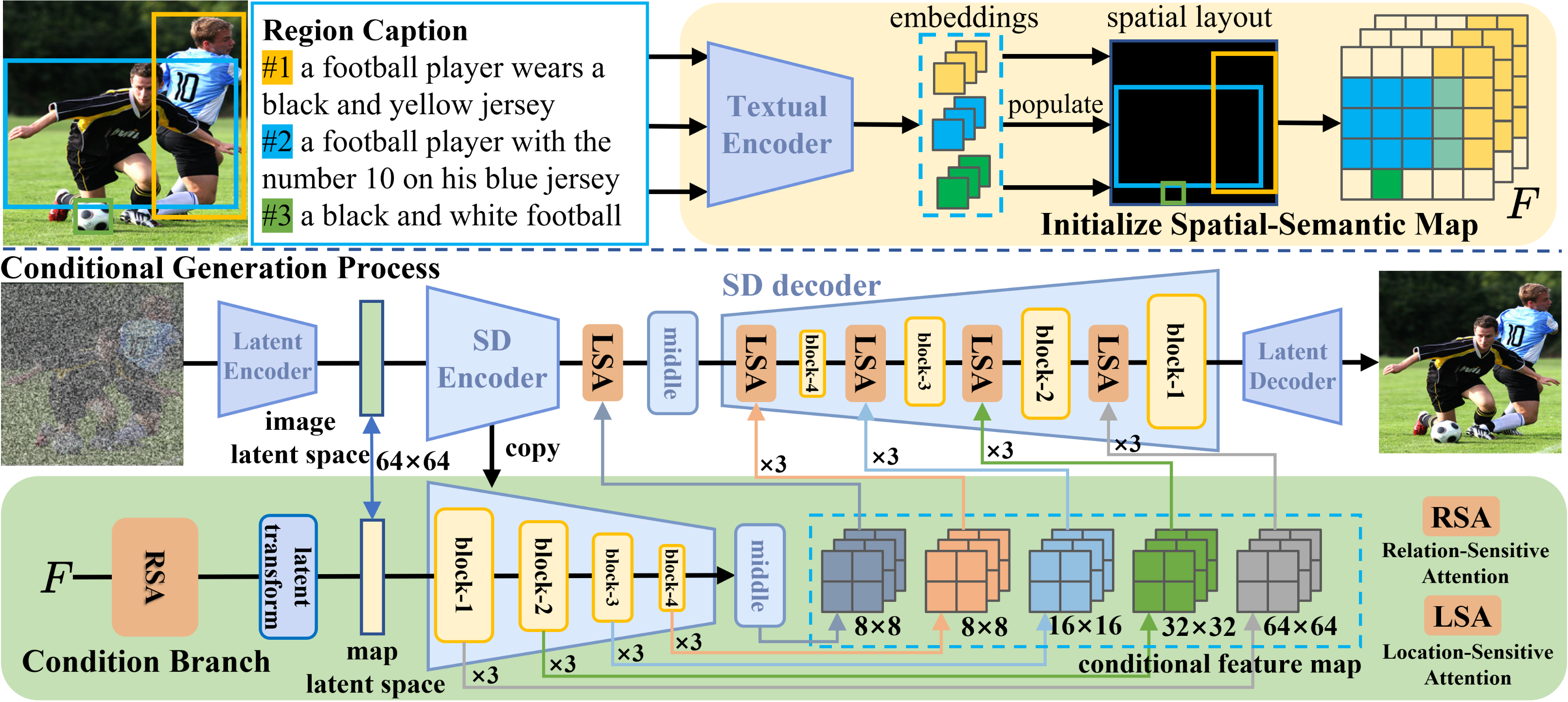}
  \caption{The overall architecture of the proposed SSMG. During the conditional generation process, we first leverage the VQ-GAN's latent encoder within Stable Diffusion (SD) to downsize the entire dataset of 512 × 512 images into the 64 × 64 latent space. To ensure consistency, we also transform the spatial-semantic map to the 64 × 64 latent space in the condition branch. Subsequently, we duplicate the structures and weights of the SD encoder and middle block as ControlNet.
  The latent map is then fed into the copied SD encoder and middle block to produce conditional feature maps at different scales. These conditional feature maps are then integrated with the corresponding blocks in the SD decoder and middle block through the proposed LSA.
  }
  \label{fig:overview}
\end{figure*}

In this paper, we present an extension to the traditional L2I generation by introducing a novel method that renders both the textual description $t_k$ and positional information $p_k$ in a free-form manner. 
For $t_k$, traditional methods have typically restricted it to a predefined set of class descriptions, \eg, ``person'' or ``ball''. Our method, however, broadens this definition to accommodate free-form long text, thereby offering more detailed semantic control over each instance. 
For example, instead of a generic ``person'' in Figure \ref{fig:overview}, we could specify ``a football player with the number 10 on his blue jersey''. What's more, our free-form L2I generation allows $p_k$ to be represented in various forms of layouts, \eg, boxes, masks, or keypoints. 
This flexibility breaks the constraints of previous methods that were limited to handling a single type of layout and thus makes it more convenient for various applications, \eg, image-to-image translation \cite{wang2022pretraining}, person image synthesis \cite{men2020controllable}.

\subsection{Map-guided Diffusion Model}
In this section, we present our Spatial-Semantic Map Guided (SSMG) diffusion model in detail. An overview of SSMG is illustrated in Figure \ref{fig:overview}. Our approach is characterized by three primary parts: (a) Initialize the spatial-semantic map that is based on the given positional information and textual descriptions. (b) Relation-Sensitive Attention (RSA) module that enhances the initial spatial-semantic map by modeling relationships among different instances and the overall scene. (c) Location-Sensitive Attention (LSA) module that enables conditional generation based on the designed spatial-semantic map, achieving sufficient integration of layout information while striving to retain the capabilities of the large-scale pre-trained T2I model. 
In the following sections, we will further elaborate on each of these steps and modules, shedding light on their contributions to our overall model.

\paragraph{Initialize Spatial-Semantic Map} 
Our objective here is to generate a 2D feature map that is rich in both semantic information and spatial location information to serve as a guiding signal. As depicted in Figure \ref{fig:overview}, to capture semantic information for each instance, we leverage rich semantics learned in the pre-trained text-to-image model by inputting each instance's description into its textual encoder. 
Considering that the related layout dataset does not provide corresponding descriptions for each instance, we crop each instance according to its region. The cropped instance is then fed into a pre-trained caption model to obtain a corresponding region caption, which serves as its instance description.

Subsequently, for each pixel $(i,j)$ associated with a specific instance $e_k=(t_k,p_k)$, we populate the corresponding position $(i,j)$ in the feature map with the extracted semantics of that instance's $t_k$. We use $f_{text}(t_k) \in \mathbb{R}^C$ to represent the pre-trained textual encoder and the initialization process of feature map $\boldsymbol{F} \in \mathbb{R}^{H \times W \times C}$ is specified as 
\begin{align*}
\label{equ:backbone}
\boldsymbol{F}(i,j,:) = f_{text}(t_k), \; \; \text{if} \; (i,j) \; \text{in } \;  p_k,
\end{align*}
where $t_k,p_k$ refers to the textual description and positional information of instance $e_k$. It's noteworthy that if a pixel $(i,j)$ does not belong to any instance, the feature at this position remains unpopulated. In cases where a pixel is associated with multiple instances (due to overlap), the value filled in is the average of the embeddings of the multiple instances.

The above procedure not only guarantees the utmost preservation of positional information by fully inheriting the spatial structure from the layout but also integrates rich semantics into the feature map. Furthermore, this initialized way can be easily extended to other types of layouts. All that is required is to populate the corresponding semantics based on the new positional information $p_k$. 

\paragraph{Relation-Sensitive Attention}
Evidently, the initial spatial-semantic map processes the semantic and spatial information of each instance independently, while neglecting the relationships among instances in the scene, as well as the relationship of each instance with the overall scene. 
This brings that some generated objects are incongruous with other objects or the entire scene, often appearing inappropriate especially when facing scenes with multiple objects or a high degree of overlap (as seen case (c) and (d) in Figure \ref{fig:case} ). 
To overcome this limitation, we propose the Relation-Sensitive Attention (RSA) module. 
This module allows the spatial and semantic information of each instance to cross-reference all other instance or scene information, thereby integrating contextual information into the feature representation of each instance. 

Concretely, we first introduce an auxiliary scene token $\boldsymbol{g}=f_{text}(t_{global})$ to represent the global scene information. Then, inspired by \cite{yang2022modeling}, we modified vanilla self-attention with a relation matrix $\boldsymbol{M}$ as follows:
\begin{align*}
\operatorname{Attention}(Q, K, V, \boldsymbol{M})=\operatorname{softmax}\left(\frac{Q K^{T}}{\sqrt{d}} \circ \boldsymbol{M}\right) V,   
\end{align*}
where $\circ$ indexes element-wise product and $d$ is the dimension of queries and keys. The attention is performed over the concatenation of flattened feature map $\boldsymbol{F}$ and scene token $\boldsymbol{g}$ where $Q=\varphi_{Q}([\boldsymbol{F},\boldsymbol{g}]), K=\varphi_{K}([\boldsymbol{F},\boldsymbol{g}]), V = \varphi_{V}([\boldsymbol{F},\boldsymbol{g}])$, and $\varphi_{Q},\varphi_{K},\varphi_{V}$ are linear projection layers. We then construct the relation matrix $\boldsymbol{M}$ in two ways, \ie, instance-instance $R_{\text {inst}}$, and instance-scene $R_{\text {scene}}$, such that
\begin{align*}
\boldsymbol{M}[s_m,s_n]=\left\{\begin{array}{ll}
1, & \text { if } R_{\text {inst}}(s_m, s_n) \text{ or } R_{\text {scene}}(s_m, s_n) \\
-\infty, & \text { else }.
\end{array}\right.
\end{align*}

\begin{figure}[t]
  \centering
  \includegraphics[width=0.48\textwidth]{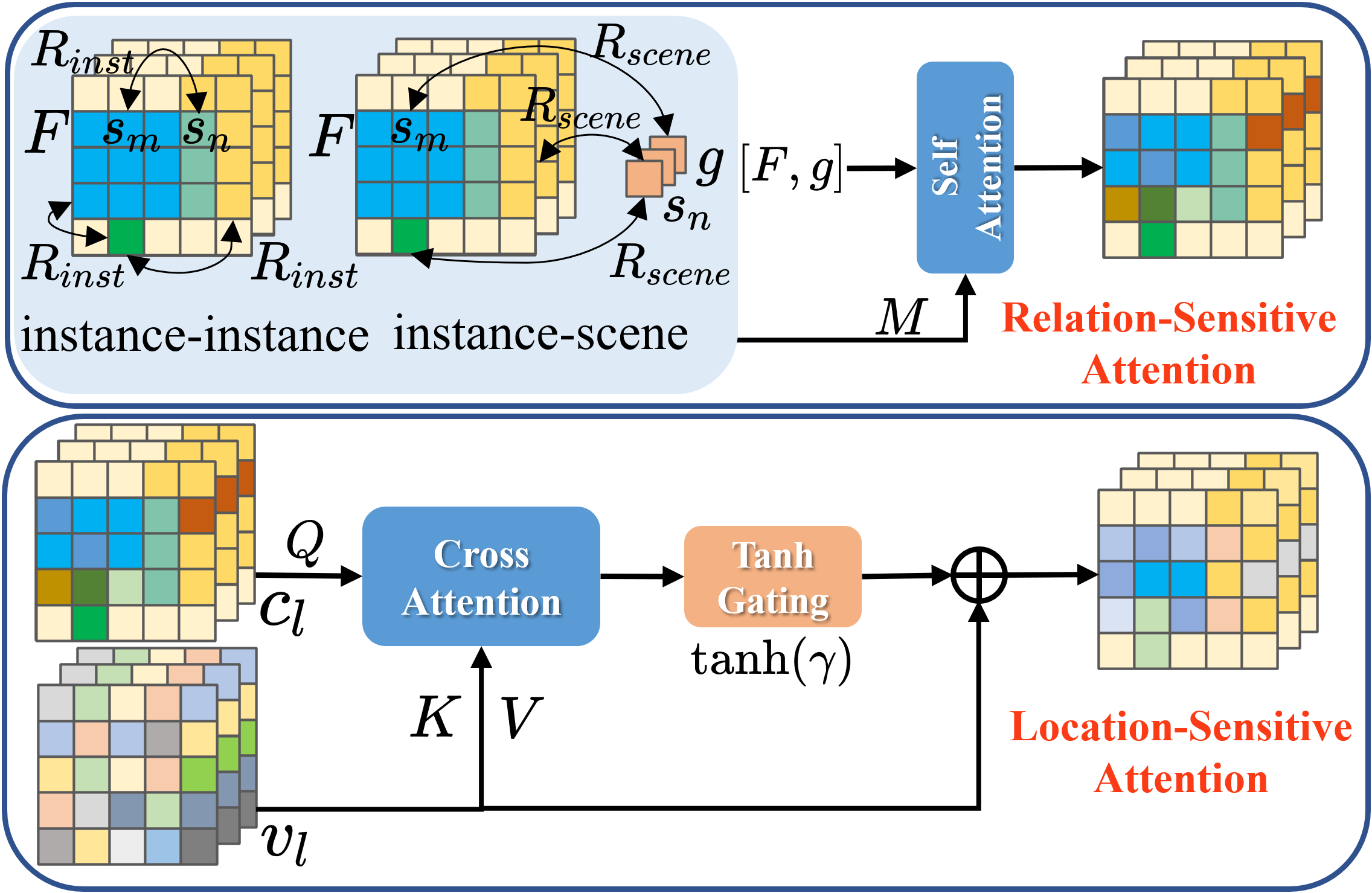}
  \caption{Illustrations of the RSA and LSA mechanisms. Feature maps are flattened before being fed into attention.}
  \label{fig:attention}
\end{figure}

As shown in Figure \ref{fig:attention}, we categorize two pixels (or tokens) $s_m$ and $s_n$ as instance-instance relative $R_{\text {inst}}(s_m,s_n)$, if they belong to distinct entities. In the case of overlapping, we prioritize treating them as belonging to different entities, thereby emphasizing the distinction and relationship between instances. We then define two pixels (or tokens) as instance-scene relative $R_{\text {scene}}(s_m,s_n)$, when one token corresponds to the scene token and the other is associated with an instance entity. The construction of $\boldsymbol{M}$ ensures that the relationships between instances, and their relationship with the overall scene, are adequately considered. Following the attention operation, the scene token is discarded, and the feature map reverts to its original shape. 

\paragraph{Location-Sensitive Attention}
In the subsequent stage, we focus on integrating the previously generated condition (\ie, enhanced spatial-semantic map) into the conditional generation process. Note that previous methods, such as channel-wise concatenation in Palette \cite{wang2022pretraining} or feature addition in ControlNet \cite{cite}, encounter severe pixel-wise structural bias due to the randomness of noise features \cite{Zhu_2023_CVPR_tryondiffusion}. Such bias notably undermines the noise feature's sensitivity to positional information within the conditions. Therefore, we introduce the Location-Sensitive Attention (LSA) to warp noise features according to the spatial-semantic map for avoiding bias. This mechanism is shown in Figure \ref{fig:attention}, formulated by
\begin{align*}
 v_l=v_l+ \tanh (\gamma) \cdot \operatorname{Attention}\left(\varphi_{Q}(c_l), \varphi_{K}(v_l), \varphi_{V}(v_l)\right),   
\end{align*}
where $v_l$ and $c_l$ denote flattened noise features and spatial-semantic map at block $l$, respectively. It's worth noting that due to the scale consistency between $v_l$ and $c_l$, LSA allows us to use spatial-semantic maps as queries and noise features as keys and values. In this way, the cross-attention can leverage similarities computed by $\varphi_{Q}(c_l)*\varphi_{K}(v_l)^T$ to re-weight noise features. This provides a learnable way to represent correspondence between noise features and positional information in $c_l$, effectively enhancing the generated noise feature's location sensitivity to spatial-semantic maps. 
Further drawing inspiration from \cite{alayrac2022flamingo}, we also employ the gating mechanism by applying $\tanh (\gamma)$, where $\gamma$ is a learnable scalar initialized to 0. This initialization ensures that the training starts from the pre-trained state, thereby preserving the integrity of the large-scale diffusion model and enhancing the stability of the training.


\subsubsection{Model Fine-tuning} 
We fine-tune the Stable Diffusion with the same LDM objective \cite{rombach2022high}, which is based on the layout information of instance entities $\boldsymbol{e}$, \ie,
\begin{align*}
 \min _{\theta} \mathcal{L}_{\text {}}=\mathbb{E}_{z,t^{\prime},\epsilon \sim \mathcal{N}(0, I)} \left[\left\|\epsilon-f_{\theta}\left(z_{t^{\prime}}, t^{\prime}, f_{text}(t_{global}), \boldsymbol{e}\right)\right\|_{2}^{2}\right],   
\end{align*}
where $z$ represents the latent code extracted from the input image; $t^{\prime}$ refers to the time step. We retain the text-guided signal $f_{text}(t_{global})$ in LDM, which also serves as the scene token $\boldsymbol{g}$. SSMG exclusively focuses on fine-tuning the conditional U-Net $f_{\theta}$ while keeping the text encoder and the SD model's latent encoder and decoder frozen during training.

\section{Experiments}

\subsection{Experimental Setup}

\subsubsection{Datasets.} We adopt widely recognized benchmarks COCO-Thing-Stuff \cite{lin2014microsoft,46624} for both training and evaluation. It consists of $118,287$ training and $5,000$ validation images, which are annotated with $80$ thing/object classes and $182$ semantic stuff classes. Following \cite{LAMA}, we disregard objects that occupy less than $2\%$ of the entire image and only utilize images containing between 3 to 8 objects. 



\subsubsection{Evaluation Metrics.} We adopt two widely recognized metrics, Inception Score (IS) \cite{is} and Fréchet Inception Distance (FID) \cite{fid} to evaluate the fidelity of the generated images. Additionally, to measure the diversity, we compute the Diversity Score (DS) between two images generated from the same layout by comparing the LPIPS \cite{LPIPS} metric in a deep neural network feature space. 
In line with metrics in previous studies, we adopt \textit{YOLO score} mAP/mAP50/mAP75 to evaluate grounding alignment and semantic accuracy between the generated images and layouts. 



\subsubsection{Implementation Details.} Our model is implemented based on the Stable Diffusion and ControlNet. During training, we take the AdamW as the optimizer within the PyTorch Lightning framework. We resize the input images to $512\times512$. The model is trained on 4 NVIDIA-A100 GPUs with a batch size of 64, requiring $\sim$ 2 days for 50 epochs. During inference, we use 20 DDIM \cite{songdenoising} sampling steps with classifier-free guidance \cite{ho2022classifier} scale of 9. For qualitative and quantitative evaluations on COCO, we use its caption annotations as $t_{global}$. For the free-form generation, we concatenate the text descriptions of each sub-instance to serve as $t_{global}$. 

\begin{figure*}[t]
  \centering \includegraphics[width=0.92\textwidth]{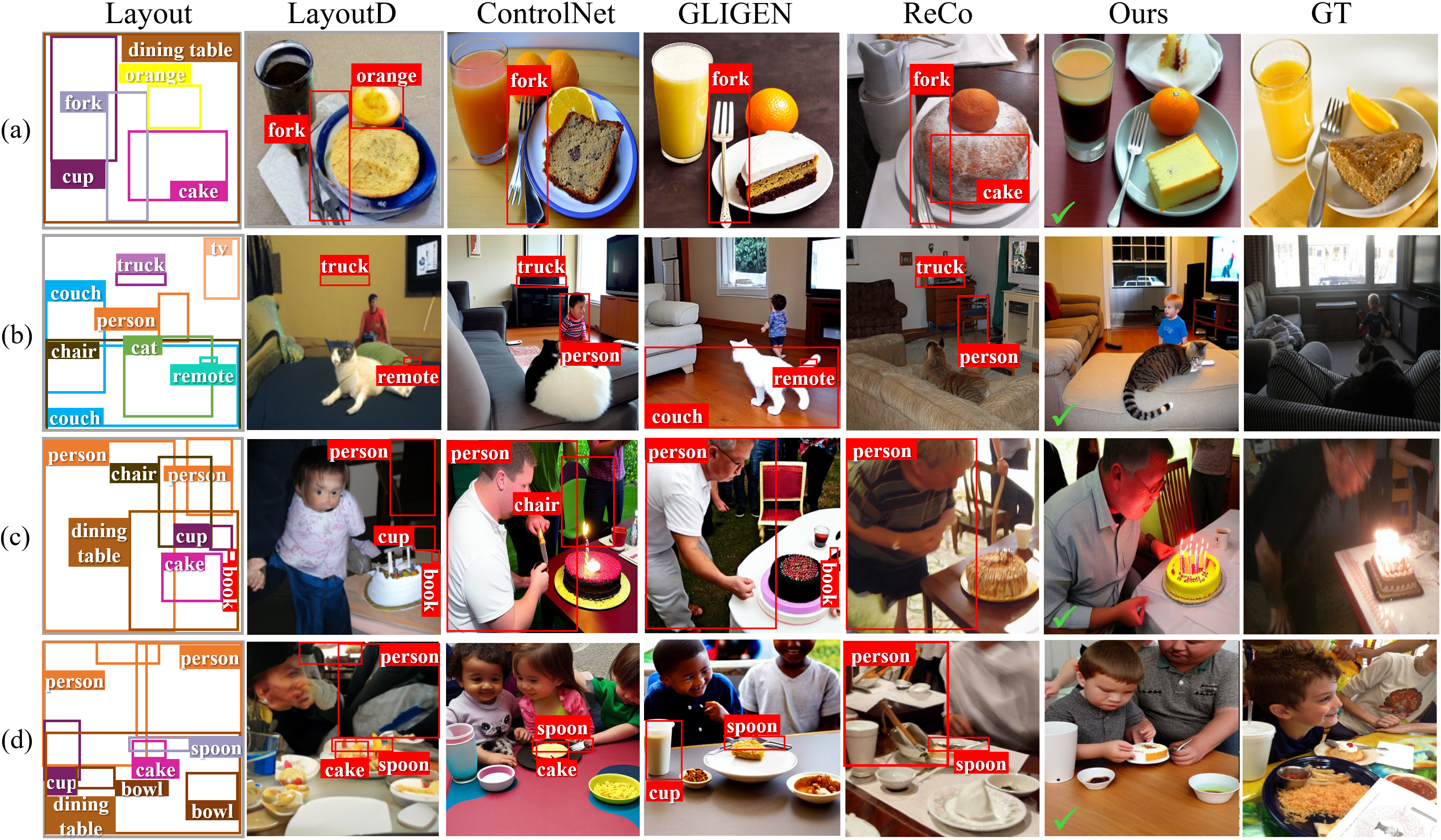}
  \caption{Qualitative comparison with SOTA methods. The red boxes indicate unrecognizable or mispositioned instances.}
  \label{fig:case}
\end{figure*}
\subsection{Quantitative Comparison}

We benchmark our method against the state-of-the-art L2I methods with bounding boxes including LostGAN \cite{sun2019image}, Con L2I \cite{he2021context}, LAMA \cite{LAMA}, TwFA \cite{yang2022modeling}, ReCo \cite{yang2023reco}, GLIGEN \cite{li2023gligen}, LayoutD \cite{zheng2023layoutdiffusion}. We provide two versions built upon Stable Diffusion v15 and v21 \cite{patil2022stable}, respectively. 
As demonstrated in Table \ref{tab:benchmark_box}, the proposed model outperforms all the competitors across FID, IS, and DS metrics, which attests to the high visual quality and diversity of the images synthesized by our method. It's noteworthy that SSMG exhibits a relatively minor improvement compared to GLIGEN and ReCO, as indicated by the FID (20.82 \vs 21.04) and IS (32.18 \vs 31.63). 
This is anticipated, given that the generative capabilities of these models are largely underpinned by the large-scale T2I models.

Crucially, our method demonstrates superior controllability compared to these methods with similar infrastructure. The YOLO score, a critical metric in our evaluation, is significantly higher for our method compared to the others. 
This metric not only reflects the model's spatial controllability over localization accuracy but also its semantic controllability to generate fine-grained details at specific locations. 
These results underscore the limitations of token-guided and image-guided methods as we discussed. Significantly, they highlight the effectiveness of our map-guided strategy in enhancing both spatial and semantic controllability. 



\begin{table}[!t]
  \tabcolsep=5 pt	
  \centering
  
  \begin{tabular}{lcccc}
  \toprule
  \textbf{Methods} & \textbf{FID$\downarrow$} & \textbf{IS$\uparrow$} & \textbf{DS $\uparrow$} & \textbf{YOLO score$\uparrow$}\\
  \cmidrule(lr){1-1} \cmidrule(lr){2-2} \cmidrule(lr){3-3} \cmidrule(lr){4-4} \cmidrule(lr){5-5} 
  LostGAN & 42.55 & 18.01\small{$\pm$0.50} & 0.45 & 9.1/15.3/9.8 \\
  Con L2I & 29.56 & 18.57\small{$\pm$0.54} & 0.65 & 10.0/14.9/11.1 \\
  LAMA & 31.12 & 14.32\small{$\pm$0.58} & 0.48 & 13.4/19.7/14.9 \\
  TwFA & 22.15 & 24.25\small{$\pm$1.04} & 0.67 & -/28.2/20.1 \\

  LayoutD* & 22.65 & 26.73\small{$\pm$0.92} & 0.57 & 18.1/31.0/18.9 \\ 
  GLIGEN & 21.04 & - & - & 22.4/36.5/24.1 \\
  Control* & 28.41 & 28.85\small{$\pm$0.85} & 0.65 & 25.2/46.7/22.7 \\ 
  ReCo* & 27.47 & 31.63\small{$\pm$0.72} & 0.62 & 30.5/56.3/29.9 \\    
  
  \midrule
    
  Our-v15 & 22.31 & \textbf{33.99\small{$\pm$1.47}} & \textbf{0.69} & 35.9/57.0/38.5 \\
  Our-v21 & \textbf{20.82} & 32.18\small{$\pm$0.85} & 0.68 & \textbf{37.6/59.0/40.9} \\
  
  \bottomrule
  \end{tabular}
  \caption{Quantitative comparison with state-of-the-art layout-to-image methods. `$\uparrow$' stands for higher the better, `$\downarrow$' stands for lower the better. All generated images are evaluated under $256\times256$ resolution for FID, IS, DS and $512\times512$ for YOLO score. Methods marked with `*' are re-evaluated using images generated from their official code.}
  \label{tab:benchmark_box}
\end{table}


\begin{figure}[t]
  \centering
  \includegraphics[width=0.35\textwidth]{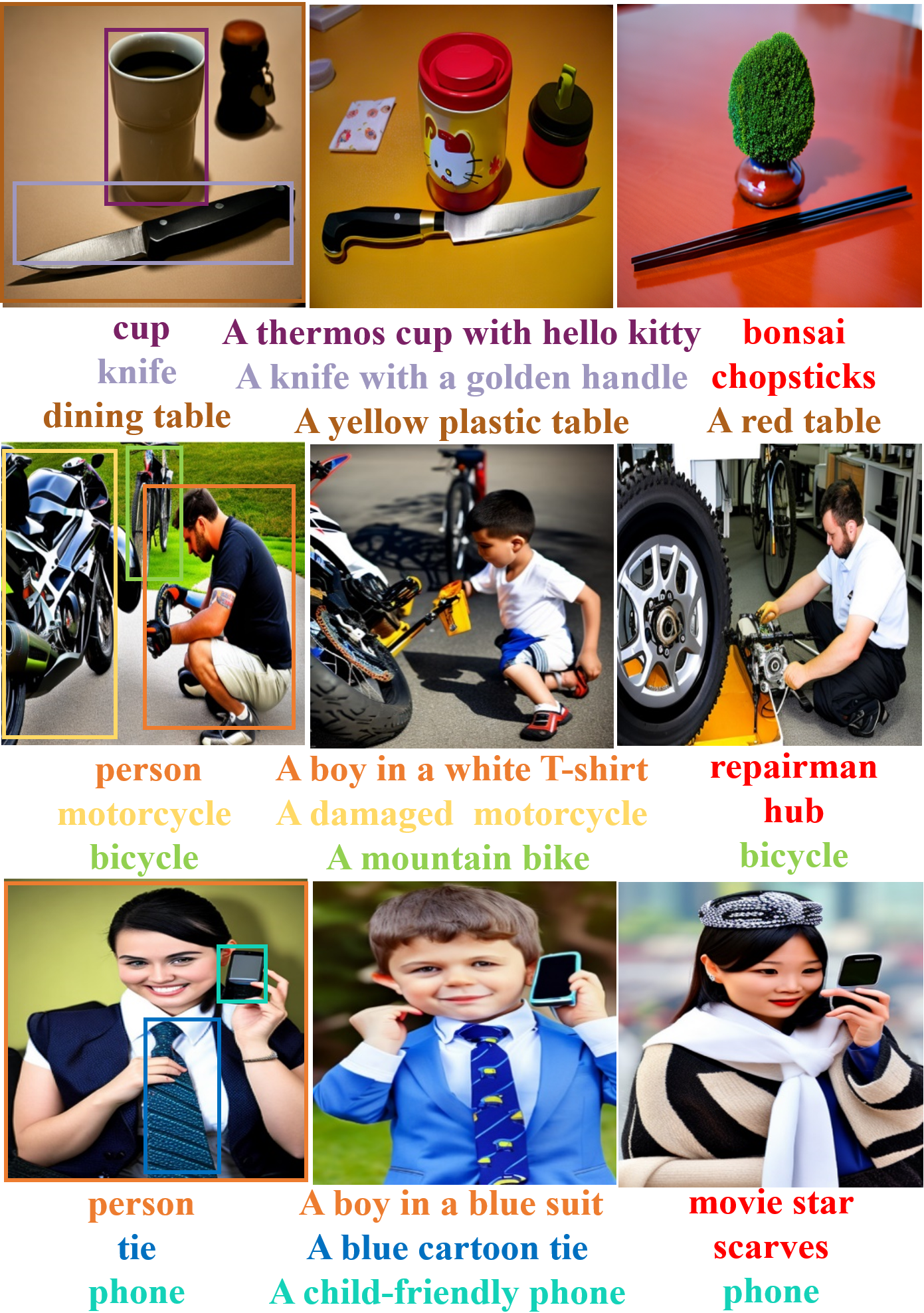}
  \caption{Illustrations of free-form textual descriptions. 
  }
  \label{fig:free_text}
\end{figure}

\begin{figure}[t]
  \centering  \includegraphics[width=0.48\textwidth]{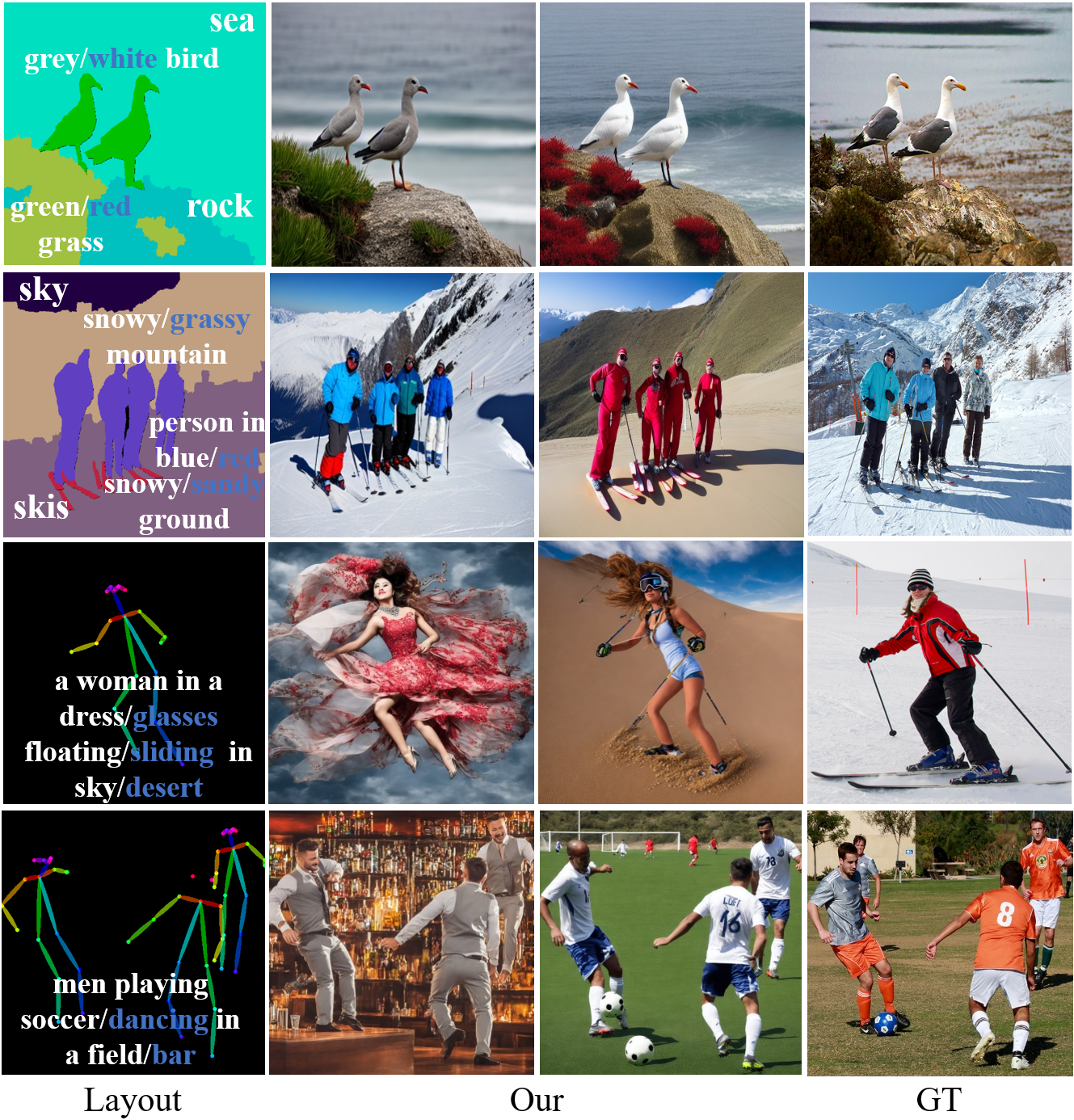}
  \caption{Illustrations of free-form positional layouts. 
  }
  \label{fig:free_layout}
\end{figure}

\subsection{Qualitative Results}
\subsubsection{Comparison} Following our quantitative comparison, we further delve into a qualitative analysis of SSMG, presenting representative case studies to highlight the distinctive features of our approach. 
As shown in Figure \ref{fig:case} (a) and (b), our method not only preserves the overall realism  of generated images but also ensures the fidelity of each individual instance. 
This stands in stark contrast to methods LayoutD \cite{zheng2023layoutdiffusion} and ControlNet, which exhibit distorted and unrealistic phenomena on local objects, despite utilizing similar network architectures. Meanwhile, state-of-the-art methods like GLIGEN \cite{li2023gligen} and ReCo \cite{yang2023reco}, although producing visually appealing images at first glance, continue to grapple with the challenges inherent to token-guided strategies, including imprecise boundaries and loss of objects. Our method, on the other hand, leverages the spatial control strength of our map-guided strategy and the Location-Sensitive Attention (LSA) module, demonstrating precise localization capabilities. As evidenced by all the cases in Figure \ref{fig:case}, the objects generated by our method are well-confined within the layout, exhibiting an exceptional sense of spatial accuracy.

Especially when dealing with a set of multiple objects with complex relationships, \eg, case (c) and (d), previous methods often struggle to generate recognizable objects at overlapping positions. However, our method, benefiting from the relationship modeling of Relation-Sensitive Attention (RSA), is capable of generating scene images that maintain complex object relationships. The overlap and interaction between objects within the image are well represented, demonstrating the model's nuanced scene understanding.

\subsubsection{Free-form L2I}
Our approach can serve as an enhancement to traditional T2I methods, permitting free-form textual descriptions and a multitude of layout positional representations. As shown in Figure \ref{fig:free_text}, SSMG is capable of leveraging free-form descriptions for each instance, allowing for the specific generation of particular styles or characteristics, \eg, from ``cup'' to ``a thermos cup with hello kitty''. Moreover, training and inference with the free-form textual descriptions enable our method to inherit the zero-shot capabilities of T2I models. As illustrated in Figure \ref{fig:free_text}, even though fine-tuned on COCO, our method can achieve out-of-distribution generation. For instance, the right column indicates that our method can generate novel content, \eg, ``bonsai'', ``chopsticks'', and ``scarves'', which falls outside the scope of the COCO dataset. 

We further demonstrate that SSMG can be applied to various positional representations, \ie, masks and keypoints. 
Figure \ref{fig:free_layout} demonstrates that our method can not only accurately generate instances at corresponding positions according to other layout forms but also control the diverse styles of instances through textual descriptions.
This remarkable combination of controllability and flexibility significantly elevates the practical value of generative models, positioning L2I generation as a potent tool for various applications. 


\subsection{Ablation studies}
To further validate the effectiveness of our proposed components, we conduct a series of ablation studies. These studies focus on the three key components of our model: the Map-Guided (MG) strategy, the mechanisms of Relation-Sensitive Attention (RSA) and Location-Sensitive Attention (LSA). We build the baseline based on image-guided ControlNet and add three components sequentially. 
As illustrated in Table \ref{tab:ablation_study}, substituting the image-guided strategy with our map-guided strategy results in a noticeable enhancement in both the fidelity and grounding accuracy of the generated images. Similarly, the addition of the RSA and LSA mechanisms also results in a significant improvement. With the RSA mechanism, the model excels at accurately capturing the relationships between different objects within the scene, resulting in higher fidelity. With the LSA mechanism, the model's spatial awareness is greatly enhanced, leading to more accurate positioning of objects.
These incremental studies clearly demonstrate the importance of each of these components to the overall performance. The map-guided strategy, coupled with the RSA and LSA, is integral to our model's ability to generate high-quality images with precise control over both fine-grained semantics and spatial layouts. 

\begin{table}[t]

 \centering
 \resizebox{0.48\textwidth}{!}{
 \begin{tabular}{@{}ccc|cc@{}}
  
  \toprule
  \textbf{MG} & \textbf{RSA} & \textbf{LSA} & \textbf{FID} $\downarrow$ & \textbf{YOLOScore} $\uparrow$ \\ 
  \midrule
        &       &       & 28.41 {(+7.59)} & 25.2/46.7/22.7 {(-14.3)} \\

  \checkmark &       &       & 25.12 {(+4.30)}  & 30.5/53.0/29.6 {(-8.1)} \\
        
  \checkmark & \checkmark &       & 21.41 {(+0.59)}  & 32.7/54.7/32.9 {(-5.7)} \\
  
  \checkmark &       & \checkmark & 23.49 {(+2.67)} & 35.1/56.8/37.5 {(-2.7)} \\
  
  \checkmark & \checkmark &  \checkmark & \textbf{20.82} & \textbf{37.6/59.0/40.9} \\

  \bottomrule
 \end{tabular}
 }
 \caption{Ablations of Map-Guided (MG), Relation-Sensitive Attention (RSA), and Location-Sensitive Attention
(LSA).}
 \label{tab:ablation_study}

\end{table}

\section{Conclusions}

This paper introduces a novel Spatially-Semantic Map Guided diffusion model that effectively addresses the limitations of previous token-guided and image-guided L2I methods. Our method, through its innovative map-guided strategy and bespoke attention mechanisms, delivers superior performance in terms of fidelity, diversity, and controllability. Furthermore, our method can be extended to be free-form, providing users with a more diverse and flexible way to describe layouts. We hope that this work will constitute an advancement in the field of layout-to-image generation, opening up new possibilities for future research and applications.


\section{Ethics Statement}
Our method's versatility opens up possibilities for its application across various domains of structured image generation. However, we must also pay attention to societal impacts that could arise from its misuse. For instance, the risk of data leakage could lead to privacy concerns and the model's capacity to produce highly realistic images could be exploited for illegal or harmful content. It's essential to follow clear usage guidelines and ensure responsible and ethical use.

\section{Acknowledgments}
This work is supported by the National Key Research and Development Program of China (No. 2022YFB3102600), National Nature Science Foundation of China (No. 62192781, No. 62272374, No. 62202367, No. 62250009, No. 62137002), Project of China Knowledge Center for Engineering Science and Technology,  Project of Chinese academy of engineering ``The Online and Offline Mixed Educational Service System for `The Belt and Road' Training in MOOC China'', and the K. C. Wong Education Foundation.

\bibliography{aaai24}

\end{document}